\documentclass[11pt,a4paper]{article}
\usepackage[hyperref]{naaclhlt2019}
\usepackage{times}
\usepackage{latexsym}
\usepackage[algo2e]{algorithm2e} 
\usepackage{graphicx}
\usepackage{graphics}
\usepackage{subcaption}
\usepackage{url}
\usepackage{booktabs}
\usepackage{multirow}
\usepackage{algorithm}
\usepackage{algorithmic}
\usepackage{placeins}
\usepackage{etoolbox}
\usepackage{makecell}
\usepackage{fancyhdr}
\usepackage{amsfonts}

\usepackage{fancyhdr}
\usepackage{geometry}
\usepackage{amsmath}
\DeclareMathOperator*{\argmin}{arg\,min}
\DeclareGraphicsExtensions{.pdf,.jpeg,.png,.jpg}

\aclfinalcopy 
\def\aclpaperid{894} 

\title{Aligning Vector-spaces with Noisy Supervised Lexicons}
  
\author{Noa Yehezkel Lubin$^\dagger$ \;\; Jacob Goldberger$^\ddagger$ \;\; Yoav Goldberg$^{\dagger *}$ \\
\texttt{noa.kel@gmail.com} \\
\\ $^\dagger$Computer Science Department, Bar Ilan University, Ramat Gan, Israel 
\\ $^\ddagger$Electrical Engineering Department, Bar Ilan University, Ramat Gan, Israel \\
$^*$Allen Institute for Artificial Intelligence
}

\date{}

\begin{document}
\thispagestyle{fancy}

\maketitle
\begin{abstract}
The problem of learning to translate between two vector spaces given a set of aligned points arises in several application areas of NLP. Current solutions assume that the lexicon which defines the alignment pairs is noise-free. We consider the case where the set of aligned points is allowed to contain an amount of noise, in the form of incorrect lexicon pairs and show that this arises in practice by analyzing the edited dictionaries after the cleaning process. We demonstrate that such noise substantially degrades the accuracy of the learned translation when using current methods. We propose a model that accounts for noisy pairs. This is achieved by introducing a generative model with a compatible iterative EM algorithm. The algorithm jointly learns the noise level in the lexicon, finds the set of noisy pairs, and learns the mapping between the spaces. We demonstrate the effectiveness of our proposed algorithm on two alignment problems: bilingual word embedding translation, and mapping between diachronic embedding spaces for recovering the semantic shifts of words across time periods.
\end{abstract}

\section{Introduction}
\label{sec:Into}

We consider the problem of mapping between points in different vector spaces.  This problem has prominent applications in natural language processing (NLP). Some examples are creating bilingual word lexicons \cite{mikolov2013exploiting}, machine translation \cite{artetxe2016learning,artetxe2017acl,artetxe2017unsupervised,artetxe2018generalizing, artetxe2018robust, conneau2017word}, hypernym generation \cite{yamane2016distributional}, diachronic embeddings alignment \cite{hamilton2016diachronic} and domain adaptation \cite{barnes2018projecting}. In all these examples one is given word embeddings in two different vector spaces, and needs to learn a mapping from one to the other.

The problem is traditionally posed as a supervised learning problem, in which we are given two sets of vectors (e.g.: word-vectors in Italian and in English) and a \emph{lexicon} mapping the points between the two sets (known word-translation pairs). Our goal is to learn a mapping that will correctly map the vectors in one space (e.g.: English word embeddings) to their known corresponding vectors in the other (e.g.: Italian word embeddings). The mapping will then be used to translate vectors for which the correspondence is unknown. This setup was popularized by \citet{mikolov2013exploiting}.

The supervised setup assumes a perfect lexicon. Here, we consider what happens in the presence of \emph{training noise}, where some of the lexicon's entries are incorrect in the sense that they don't reflect an optimal correspondence between the word vectors. 

\section{Background}

\subsection{The Supervised Translation Problem}
We are given two datasets, $X=x_1,...,x_m$ and $Y=y_1,...,y_n$, coming from $d$-dimensional spaces $\mathcal{X}$ and
$\mathcal{Y}$.
We assume that the spaces are related, in the sense
that there is a function $f(x)$ mapping points in space
$\mathcal{X}$ to points in space $\mathcal{Y}$. In this work, we focus on linear
mappings, i.e. a $d\times d$ matrix $Q$ mapping points via $y_i=Qx_i$. The goal of
the learning is to find the translation matrix $Q$. In the supervised setting, $m=n$ and we assume that $\forall i$  $f(x_i)\approx y_i$. We refer to the sets $X$ and $Y$ as the \emph{supervision}. The goal is to learn a matrix $\hat{Q}$ such the Frobenius norm is minimized:
\vspace{-0.2cm}
\begin{equation}
\hat{Q} = \argmin_{Q} \|QX - Y\|_F^2.
\label{eq:goal}
\end{equation}

\subsection{Existing Solution Methods}
\label{sec:existing_methods}
\paragraph{Gradient-based}
The objective in (\ref{eq:goal}) is convex, and can be solved via least-squares method or via stochastic gradient optimization iterating over the pairs $(x_i,y_i)$, as done by \citet{mikolov2013exploiting} and \citet{DBLP:journals/corr/DinuB14}.

\paragraph{Orthogonal Procrustes (OP)}
\label{par:procrustes}
\citet{artetxe2016learning} and \citet{DBLP:journals/corr/SmithTHH17} argued and proved that a linear mapping between sub-spaces must be orthogonal. This leads to the modified objective:
\vspace{-0.2cm}
\begin{equation}
 \hat{Q} = \argmin_{Q,  s.t: Q^TQ = I} \|QX - Y\|_F^2
 \label{eq:goal-orth}
\end{equation}

Objective (\ref{eq:goal-orth}) is known as the \emph{Orthogonal Procrustes Problem}. It can be solved algebraically 
by using a singular value decomposition (SVD).  \citet{procrustes1966} proved
that the solution to ~\ref{eq:goal-orth}  is: $\hat{Q} = UV^T$ s.t. $ U\Sigma V^T $  is the SVD of $YX^T$.

The OP method is used in \citet{xing2015normalized, artetxe2016learning, artetxe2017acl, artetxe2017unsupervised, artetxe2018generalizing, artetxe2018robust, hamilton2016diachronic, conneau2017word, ruder2018discriminative}.

\subsection{The Unsupervised Translation Problem}
The supervised alignment problem can be expended to the semi-supervised \cite{artetxe2017unsupervised, lample2017unsupervised, ruder2018discriminative} or unsupervised \cite{zhang2017adversarial, conneau2017word, artetxe2018robust, xu2018unsupervised, alvarez2018gromov} case, where a very small lexicon or none at all is given. In iterative methods, the lexicon is expended and used to learn the alignment, later the alignment is used to predict the lexicon for the next iteration and so on. In adversarial methods, a final iterative step is used after the lexicon is built to refine the result. We will focus on the supervised stage in the unsupervised setting, meaning estimating the alignment once a lexicon is induced.

\begin{figure}
  \centering
  \includegraphics[height=8cm]{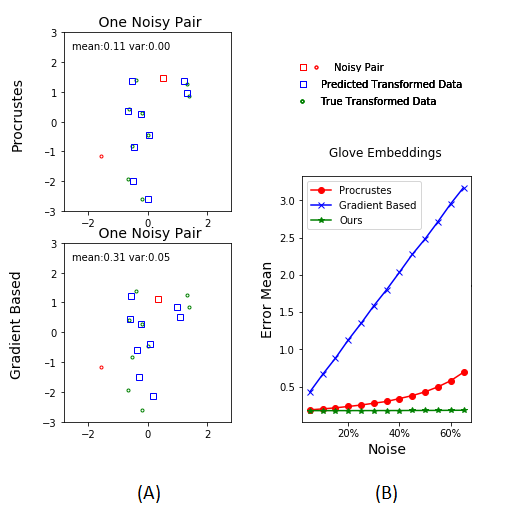}
  \caption{Noise influence.  (A): the effect of a noisy pair on 2D alignment. (B) mean error over non-noisy pairs as a function of noise level.}
    \label{fig:2d}
\end{figure}

\section{The Effect of Noise}
\label{sec:noise_effect}
The previous methods assume the supervision set $X,Y$ is perfectly correct. However, this is often not the case in practice. We consider the case where a percentage $p$ of the pairs in the supervision set are ``noisy": applying the gold transformation to a noisy point $x_j$ will not result in a vector close to $y_j$.
The importance of the quality of word-pairs selection was previously analyzed by \citet{vulic2016role}. Here, we equate ``bad pairs" to noise, and explore the performance in the presence of noise by conducting a series of synthetic experiments. We take a set of points $X$, a random transformation $Q$ and a gold set $Y=QX$. We define \emph{error} as $\|Y-\hat{Y}\|^2_F$ where $\hat{Y} = \hat{Q}X$ is the prediction according to the learned transform $\hat{Q}$. Following the claim that linear transformations between word vector spaces are orthogonal, we focus here on orthogonal transformations. 

\paragraph{Low Dimensional Synthetic Data}

\label{par:2d_setup}
We begin by inspecting a case of few 2-dimensional points, which can be easily visualized.
We compare a noise-free training to the case of a single noisy point.
We construct X by sampling \(n=10\) points of dimension \(d=2\) from a normal distribution. We take nine points and transformed them via an orthogonal random transform \(Q\). We then add a single noisy pair which is generated by sampling two normally distributed random points and treating them as a pair. The error is measured only on the nine aligned pairs.

When no noise is applied, both Gradient-based and Procrustes methods are aligned with 0 error mean and variance. Once the noisy condition is applied this is no longer the case. Figure~\ref{fig:2d}(A) shows the noisy condition. Here, the red point (true) and box (prediction) represent the noisy point. Green dots are the true locations after transformation, and the blue boxes are the predicted ones after transformation. Both methods are affected by the noisy sample: all ten points fall away from their true location. The effect is especially severe for the gradient-based methods.

\paragraph{High Dimensional Embeddings} The experiment setup is as before, but instead of a normal distribution we use (6B, 300d) English Glove Embeddings \cite{pennington2014glove} with lexicon of size \(n = 5000\). We report the mean error for various noise levels on an unseen aligned test set of size 1500.

\FloatBarrier
\begin{table*}
\centering
\resizebox{\textwidth}{!}{\begin{tabular}{| *{13}{c|} }
    \hline
Method & \multicolumn{3}{c|}{En\(\rightarrow\)It} & \multicolumn{3}{c|}{En\(\rightarrow\)De} & \multicolumn{3}{c|}{En\(\rightarrow\)Fi} & \multicolumn{3}{c|}{En\(\rightarrow\)Es} \\
    \hline
    & best & avg & iters & best & avg & iters & best & avg & iters & best & avg & iters \\
    \hline
\citealp{artetxe2018robust}   &   \textbf{48.53}  &   48.13  & 573 &   48.47  &   48.19  & 773&   33.50  &   32.63 &988&   37.60  &   37.33 &  808\\
    \hline
Noise-aware Alignment &  \textbf{48.53}    & \textbf{48.20} & 471&  \textbf{49.67}  &  \textbf{48.89}  & 568 &  \textbf{33.98}    &  \textbf{33.68}    & 502 &   \textbf{38.40}    &  \textbf{37.79}   &  551 \\
    \hline
\end{tabular}}
\caption{\textbf{Bilingual Experiment P@1.} Numbers are based on 10 runs of each method. The En$\rightarrow$De, En$\rightarrow$Fi and En$\rightarrow$Es improvements are significant at $p<0.05$ according to ANOVA on the different runs.}
\label{tbl:bilingual}
\end{table*}

In Figure ~\ref{fig:2d}(B) we can see that both methods are effected by noise. As expected, as the amount of noise increases the error on the test set increases. We can again see that the effect is worse with gradient-based methods.

\section{Noise-aware Model}
\label{sec:model}

Having verified that noise in the supervision severely influences the solution of both methods, we turn to proposing a noise-aware model.

The proposed model jointly identifies noisy pairs in the supervision set and learns a translation which ignores the noisy points. Identifying the point helps to clean the underlying lexicon (dictionary) that created the supervision. In addition, by removing those points our model learns a better translation matrix. 

\paragraph{Generative Model}

We are given $x \in \mathbb{R}^d$ and we sample a corresponding $y \in \mathbb{R}^d$ by first sampling a Bernoulli random variable with probability $\alpha$:
\vspace{-0.2cm}
\[ z \sim \text{Bernoulli}(\alpha)
\]
\vspace{-0.5cm}
\[
y \sim \begin{cases}
N(\mu{_y}, \sigma_y^2I) & z=0 \, \text{(`noise')}\\
N(Qx, \sigma^2I) & z=1 \, \text{(`aligned')}\\
\end{cases}
\]

\noindent
The density function  $y$  is a mixture of two Gaussians:
$$f(y|x) = (1 \!-\! \alpha) N(\mu{_y}, \sigma{_y}^2I) + \alpha N({Q}{x}, \sigma^2I).$$
The likelihood function is:
$$L(Q, \sigma, \mu{_y}, \sigma{_y}) =  \sum_{t} \log  f(y_t|x_t)$$

\paragraph{EM Algorithm}
We apply the EM algorithm \cite{dempster1977maximum} to maximize the objective in the presence of latent variables. The algorithm has both soft and hard decision variants. We used the hard decision one which we find more natural, and note that the posterior probability of $z_t$ was close to 0 or 1 also in the soft-decision case. 

It is important to properly initialize the EM algorithm to avoid convergence to a local optima. 
We initialize ${Q}$ by applying OP on the entire lexicon (not just the clean pairs). We initialize the variance, $\sigma$, by calculating \(\sigma^2 = \frac{1}{n\cdot{d}} \sum_{t=1}\|Qx_t - y_t\|^2\) . 
We initialize, $\mu{_y}$, $\sigma{_y}$ by taking the mean and variance of the entire dataset. Finally, we initialize $\alpha$ to $0.5$. 

The (hard version) EM algorithm is shown in Algorithm box 1. 
The runtime of each iteration is dominated by the OP algorithm (matrix multiplication and SVD on a $d \times d$ matrix). Each iteration contains an additional matrix multiplication and few simple vector operations.
Figure \ref{fig:2d}(B) shows it obtains perfect results on the simulated noisy data.

\begin{algorithm}[H]
 \caption{Noise-aware Alignment}

 \KwData{List of paired vectors: $(x_1,y_1),...,(x_n,y_n)$}
 
 \KwResult{$Q$, $\sigma$, $\mu{_y}$, $\sigma{_y}$}
\While{$\left|\alpha_{curr} - \alpha_{prev}\right| > \epsilon$}{
  \textbf{E step:} \;\\
  $w_t = p(z_t\! =\! 1 | x_t, y_t ) = \frac{\alpha N({Q}{x_t}, \sigma^2I)}{f(y_t|x_t)}$\\  
 $h_t = 1 (w_t>0.5)$\\
 $n_1= \sum_t h_t$  \
 
  \textbf{M step:}\\
  Apply OP on the subset $\{ t|h_t=1 \}$ to find Q.\\
  \; $\sigma^2 = \frac{1}{d \cdot{n_1}}$ $\sum_{t|h_t=1}\|Qx_t - y_t\|^2$\\
  $\mu{_y} = \frac{1}{(n-n_1)} \sum_{t|h_t=0} y_t$\\
  $\sigma_y^2 = \frac{1}{d{(n-n_1)}} \sum_{t|h_t=0} \|\mu{_y} - y_t\|^2$ \\
  $\alpha_{prev} = \alpha_{curr}$\\
  $\alpha_{curr} = \frac{n_1}{n}$ \
  
  }
  
\end{algorithm}

\vspace{-0.5cm}
\section{Experiments}
\label{sec:experiments}

\subsection{Bilingual Word Embedding}

\paragraph{Experiment Setup}
This experiment tests the noise-aware solution on an unsupervised translation problem. The goal is to learn the ``translation matrix'', which is a transformation matrix between two languages by building a dictionary. We can treat the unsupervised setup after retrieving a lexicon as an iterative supervised setup where some of the lexicon pairs are noisy. We assumes the unsupervised setting will contain higher amount of noise than the supervised one, especially in the first iterations. We follow the experiment setup in \citet{artetxe2018robust}. But instead of using OP for learning the translation matrix, we used our Noise-Aware Alignment (NAA), meaning we jointly learn to align and to ignore the noisy pairs. We used the En-It dataset provided by \citet{DBLP:journals/corr/DinuB14} and the extensions: En-De, En-Fi and En-Es of \citet{artetxe2018generalizing, artetxe2017unsupervised}. 

\paragraph{Experiment Results}
In Table ~\ref{tbl:bilingual} we report the best and average precision@1 scores and the average number of iterations among 10 experiments, for different language translations.
Our model improves the results in the translation tasks. In most setups our average case is better than the former best case. In addition, the noise-aware model is more stable and therefore requires fewer iterations to converge. The accuracy improvements are small but consistent, and we note that we consider them as a lower-bound on the actual improvements as the current test set comes from the same distribution of the training set, and also contains similarly noisy pairs. Using the soft-EM version results in similar results, but takes roughly 15\% more iterations to converge.

Table ~\ref{tbl:en_it} lists examples of pairs that were kept and discarded in En-It dictionary. The algorithm learned the pair (dog \(\rightarrow\) dog) is an error. Another example is the translation (good \(\rightarrow\) santo) which is a less-popular word-sense than (good \(\rightarrow\) buon / buona). When analyzing the En-It cleaned dictionary we see the percentage of potentially misleading pairs (same string, numbers and special characters)  is reduced from 12.1\% to 4.6\%.

\begin{table}[hbt!]  \centering
  \begin{tabular}{llll}
    \toprule
    English & Italian & Latent Variable \\
    \midrule
    dog     & cane & Aligned  \\
    dog     & cani & Aligned  \\
    dog     & dog  & Noise  \\
    good    & buon & Aligned  \\
    good    & buona & Aligned  \\
    good    & santo & Noise  \\
    new & new & Noise \\
    new & york & Noise \\
    new & nuove & Aligned \\
    \bottomrule
  \end{tabular}
  \caption{A sample of decisions from the noise-aware alignment on the English \(\rightarrow\)
  Italian dataset.}
  \label{tbl:en_it}
\end{table}

\subsection{Diachronic (Historical) Word Embedding}
\paragraph{Experiment Setup}
The goal is to align English word-embedding derived from texts from different time periods, in order to identify which words changed meaning over time. The assumption is that most words remained stable, and hence the supervision is derived by aligning each word to itself. This problem contains noise in the lexicon by definition. We follow the exact setup fully described in \citet{hamilton2016diachronic}, but replace the OP algorithm with our Noise-aware version \footnote{ Pre-possessing: removing proper nouns, stop words and empty embeddings. Post-processing: removing words whose frequency is below $10^{-5}$ in either years.}.
We project 1900s embeddings to 1990s embeddings vector-space. The top 10 distant word embeddings after alignment are analyzed by linguistic experts for semantic shift.

\paragraph{Experiment Results}
45.5\% of the input pairs were identified as noise. After the post processing of removing the non-frequent words as described in the experiment setup we end up with 121 noisy words. Our algorithm successfully identifies \textbf{all} the top-changing words in \citet{hamilton2016diachronic} as noise, and learns to ignore them in the alignment. In addition, we argue our method provides better alignment. Table ~\ref{tbl:historical} shows the Nearest Neighbor (NN) of a 1990s word, in the 1900s vector-space after projection. We look at the top 10 changed words in \citet{hamilton2016diachronic} and 3 unchanged words. We compare the alignment of the OP projection to the Noise-aware Alignment (NAA). For example, with our solution the word \textit{actually} whose meaning shifted from "in fact" to express emphasize or surprise, is correctly mapped to \textit{really} instead of \textit{believed}. The word \textit{gay} shifted from \textit{cheerful} to \textit{homosexual}, yet is still mapped to  \textit{gay} with NAA. This happens because the related embeddings (\textit{homosexual}, \textit{lesbian} and so on) are empty embeddings in 1900s, leaving \textit{gay} as the next-best candidate, which we argue is better than OP's \textit{society}. The words \textit{car, driver, eve} whose meaning didn't change, were incorrectly aligned with OP to \textit{cab, stepped, anniversary} instead of to themselves.

\begin{table}[hbt!]
\centering
\begin{tabular}{|l|l|l|l|}
\hline
\thead{1990s \\ Word}  & \thead{1900s NN \\ aligned \\with OP}   & \thead{1900s NN \\ aligned \\ with NAA} & \thead{Latent \\ Variable}  \\
\hline
\hline
\textbf{wanting} & need & wishing & Noise\\ \hline
\textbf{gay} & society & gay &Noise\\ \hline
\textbf{check} & give & send &Noise\\ \hline
\textbf{starting} & begin & beginning &Noise\\ \hline
\textbf{major} & general & successful &Noise\\ \hline
\textbf{actually} & believed & really &Noise\\ \hline
\underline{touching} & touched & touching &Noise\\ \hline
harry & hello & john &Noise\\ \hline
\textbf{headed} & halfway & toward &Noise\\ \hline
romance & artists & romance &Noise\\ \hline\hline
\textit{car} & cab & car &Aligned\\ \hline
\textit{driver} & stepped & driver &Aligned \\ \hline
\textit{eve} & anniversary   & eve &Aligned \\ 
\hline
\end{tabular}
\caption{\textbf{Diachronic Semantic Change Experiment.}
Upper-part: noisy pairs. \textbf{Bold:} real semantic shifts. \underline{Underlined:} global genre/discourse shifts. Unmarked: corpus artifacts. Bottom-part: clean pairs: \textit{Italics:} unchanged words, no semantic shift.}
\label{tbl:historical}
\end{table}

\section{Conclusion}
\label{sec:conc}
We introduced the problem of embedding space projection with \emph{noisy} lexicons, and showed that existing projection methods are sensitive in the presence of noise. We proposed an EM algorithm that jointly learns the projection and identifies the noisy pairs. The algorithm can be used as a drop-in replacement for the OP algorithm, and was demonstrated to improve results on two NLP tasks. We provide code at https://github.com/NoaKel/Noise-Aware-Alignment.

\section*{Acknowledgments}
The work was supported by The Israeli Science Foundation (grant number 1555/15), and by the Israeli ministry of Science, Technology and Space through the Israeli-French Maimonide Cooperation program. We also, thank Roee Aharoni for helpful discussions and suggestions.

\bibliography{naaclhlt2019}
\bibliographystyle{acl_natbib}

\end{document}